\documentclass[12pt]{article}
\usepackage[utf8]{inputenc}
\usepackage[T1]{fontenc}
\usepackage{fixltx2e}
\usepackage{graphicx}
\usepackage{longtable}
\usepackage{float}
\usepackage{wrapfig}
\usepackage{soul}
\usepackage{textcomp}
\usepackage{marvosym}
\usepackage{wasysym}
\usepackage{latexsym}
\usepackage{amssymb}
\usepackage{hyperref}
\usepackage{lmodern}
\usepackage{tabularx}

\title{Spike Synchronization Dynamics of Small-World Networks}
\author{Derek Harter\\Texas A\&M University - Commerce\\Commerce, TX  USA\\\texttt{Derek.Harter@tamuc.edu}}
\date{\today}
\hypersetup{
  pdfkeywords={},
  pdfsubject={Draft Manuscript},
  pdfcreator={Emacs Org-mode version 7.8.11}}

\begin{document}

\maketitle

\begin{abstract}

In this research report, we examine the effects of small-world network 
organization on spike synchronization dynamics in networks of
Izhikevich spiking units.  We interpolate network organizations from
regular ring lattices, through the small-world region, to random
networks, and measure global spike synchronization dynamics.  We
examine how average path length and clustering effect the dynamics of
global and neighborhood clique spike organization and propagation.  We
show that the emergence of global synchronization undergoes a phase
transition in the small-world region, between the clustering and path
length phase transitions that are known to exist.  We add additional
realistic constraints on the dynamics by introducing propagation
delays of spiking signals proportional to wiring length. The addition
of delays interferes with the ability of random networks to sustain
global synchronization, in relation to the breakdown of clustering in
the networks.  The addition of delays further enhances the finding
that small-world organization is beneficial for balancing neighborhood
synchronized waves of organization with global synchronization
dynamics.


\end{abstract}

\section{Introduction}
\label{sec-1}
\subsection{Neural Oscillations and Neural Constraints}
\label{sec-1-1}

The emergence of global synchronized behavior in biological brain
neural oscillations may be important for many types of information
processing tasks.  Global synchronized waves are thought to play roles
in multi-sensory feature binding, global information transfer,
consciousness and rhythmic motor entrainment, among other functions
\cite{engel-2001,ward-2003}.  In this work we investigate the effects
that global network architecture have on micro and macro
synchronization of spiking neuronal units.  The roles or usefulness of
synchronization are not the subject of this paper, but simply what
types of synchronization dynamics are facilitated or suppressed as
network architectures vary in their properties of how far and how long
information must travel in the network to get to its destination, and
how cohesive neuronal groups are in the information processing
network.

Some basic features of brain networks appear to be highly conserved
over different scales and types of measurement
\cite{bullmore-sporns-2009,boccaletti-2006}.  The archetypal brain
network appears to be a small-world network, with relatively short
path lengths for information to travel along and with high local
neighborhood clustering.  The former logically affects the efficiency
of global information processing, while clustering is useful in
robustness and error tolerance.  Brain networks appear to have
scale-free degree distributions, where a few hub nodes have large
numbers of connections while the majority of neural units have small
numbers of connections, both of which are compatible with clustering
and highly modular hierarchical neighborhood organization.  Brain
networks and nervous systems are highly constrained by the world they
must operate within.  Brain networks have neither unlimited space nor
energy in order to perform their crucial information processing
functions.  Brain networks must accomplish their tasks adequately
while conserving material and providing adequate energy for the
system.  Anatomical networks are sparsely connected and wiring lengths
are close to minimal, which certainly reflects the physical and energy
constraints that the brain faces in order to function.

The structure and reflected function of brain networks, seen on many
different scales, may be the result of the evolution and development
of the systems under these constraints.  Brain networks are spatially
embedded, with physical constraints, and rely on biological processes
to keep them supplied with energy.  Brains must achieve highly
efficient information processing and information transfer while
maintaining low connection and energy costs.  As Bullmore and Sporns
state \cite[p.~196]{bullmore-sporns-2009}: 

\begin{quote}
If wiring cost was exclusively prioritized the network would be close
to a regular lattice, whereas if efficiency was the only selection
criterion the network would be random.  The existence of a few
long-range anatomical connections can deliver benefits in terms of
efficiency and could arguably account for the evolution of economical
small-world properties in brain networks at all scales.
\end{quote}

The functional behavior of networks with small-world and scale free
structures have, of course, been studied in the past.  In
\cite{deng-2007} the authors develop a hierarchical organization of
sparsely connected internal neurons that form domains (or
clusters). The clusters were organized into a type of echo state
network (ESN), but instead of being a random network as in a typical
ESN, they subjected the ESN reservoir to incremental growth rules in
order for form a scale-free degree distribution of the reservoir
clusters.  They mainly applied their architecture to solving
prediction problems.  The system, in theory with its organization, can
be scaled to reflect some of the same natural characteristic of
scale-free biological neural networks, such as fast signal propagation
and coherent synchronization.  In \cite{wang-2002}, the work most
similar to the present research, the authors investigated synchronization
in small-world networks as well.  In this paper, the authors showed
with some analytic results as well as simulations, that with sufficient
weights and large enough networks, that a small-world network of
continuous-time oscillators will synchronize, even if the original 
nearest-neighbor coupled networks cannot achieve synchronization
under the same conditions.  The present research supports these basic
findings, as we also show empirically for spiking networks that
global synchronization is possible with sparsely connected networks
with a few long-range interconnections (small-world), but impossible
for regular networks.

In this research, we investigate some of the questions raised in this
view of brain networks as systems that have evolved under energy and
spatial constraints in order to optimize information processing.
Brain networks are especially good at being both flexible and
inventive, but robust both in the face of uncertainty and also in the
face of structural damage or other difficulties.  Better understanding
how and why small-world organization under such constraints evolve and
how they support robust information processing is the goal of research
such as that presented in this report.
\subsection{Small-world Networks}
\label{sec-1-2}

We have mentioned small-world network organization previously, and the
concept plays an important part in this presented research report.  In
this section we will review the Watts and Strogatz (WS) interpolation
model, and explain some of the basic features of small-world network
organization.  The idea of small-world organization had been around
for some time before the WS model formalized the concept.  Stanley
Milgram \cite{milgram-1967} showed empirically the possible existence
of small-world organization in human social networks in 1967 by
mailing letters to random people, asking them to send to someone they
know, who would in turn forward to someone they know, with the goal of
returning it back to the sender.  The chain such letters took showed a
surprisingly short number of connections needed on average to connect
any two people (the so called six-degrees of separation concept).

In 1998, WS proposed their interpolation model that demonstrated the
concept of small-world networks \cite{watts-strogatz-1998} by creating
networks that are intermediate between a regular (ring) lattice and a
random graph.  A ring lattice is simply a network of nodes organized
in a circle, where each node is connected to $k$ of its nearest
neighbors.  A random network, as the name implies, is a set of nodes
whose connections have been generated randomly, in the general case
where each connection has an equal probability $p$ of existing (see
Figure \ref{fig:ws-interpolated-networks}).  Random networks were first explored
by Erd\H{o}s and R\'enyi \cite{erdos-renyi-1960}, and the general
class of models are known as Erd\H{o}s-R\'enyi (ER) random graphs.

Small-world networks are networks that show both a relatively low
average path length, but at the same time exhibit high degrees of
local clustering.  Here average path length is easy to understand, the
minimal path length between any two nodes is the smallest distance (in
a weighted graph) or number of hops (in an unweighted graph) that you
need to travel to get from any particular starting point to any
particular destination.  The average path length (called simply $L$)
is then simply the average of all the minimal path lengths between all
possible pairs of nodes.  Clustering ($C$) is also a fairly simple
measure of a network, it measures to what degree nodes are connected
to one another.  For example, to calculate $C$, take all of the
immediate neighbors of a node.  For this set of nodes, determine the
number of possible connections between all the nodes, and the number
of actual connections that exist in the graph.  The ratio of the
actual to the possible connections in the neighborhood is the
clustering coefficient for the original node.  The average clustering
$C$ over the graph is simply the average clustering coefficient
calculated over all nodes.  See \cite{bullmore-sporns-2009} for a good
review of these and other network graph properties.

For a completely regular network, like a ring lattice where $N$ nodes
are organized in a circle, and each node is connected with its $k$
nearest neighbors along the ring, $C$ is very high, because neighbors
of any node tend to be connected with one another.  However, in a
regular ring lattice, the average path lengths are relatively long,
because there are no long-range connections, and thus you are forced
to take many small hops to get from one side of the ring to the other.
For a random network, however, the properties of $C$ and $L$ are
reversed.  In a random network, average path lengths are short,
because of the nature of the random connectivity.  However, there is
no real clustering to speak of, as the neighbors of any node are
unlikely to be connected to one another.  WS showed that, as we move
from a regular ring lattice towards a random network there is a region
of the networks that display both a beneficial low $L$ while still
retaining high $C$.  They also showed that both $L$ and $C$ perform a
rapid transition when interpolating from regular to random networks,
but these phase transition occurs at different locations in the phase
space.  This region of low path lengths but high neighborhood
clustering is known as the small-world region (see figure
\ref{fig:sim1-CLS-plt}).
\subsection{Izhikevich Spiking Networks}
\label{sec-1-3}

Using the Watts and Strogatz interpolation algorithm, we will be
generating networks that vary from regular through small-world to
random networks.  However, in this research report, we are interested
in studying the functional dynamics that such networks might support.
We will be using standard Izhikevich spiking units for this purpose
\cite{izhikevich-2003}.  Izhikevich introduced a model of spiking
units that were both computationally efficient and easy to simulate,
as well as capable of producing rich firing dynamics seen in real
biological neurons \cite{izhikevich-2004}.  The Izhikevich spiking
model has become very popular because of these reasons for simulating
spiking network dynamics.  In the original 2003 paper, Izhikevich
showed an example of synchronized behavior in a network of $N=1000$
Izhikevich units.  In this example, the network was fully connected
(every unit had a connection to every other unit).  He also had a mix
of excitatory and inhibitory units (in a ratio of 800 excitatory to
200 inhibitory units).  The model defines only two variables to
describe a neural unit: the voltage, or membrane potential of a neuron
$v$, and a membrane recovery variable $u$, with an auxiliary
after-spike reset defining a spike as occurring when the membrane
potential reaches or exceeds $30mV$.  A number of parameters may be
set in the model in order to produce the rich set of spiking pattern
dynamics mentioned.

All networks presented in this research report use Izhikevich spiking units
with network size $N=1000$ and a ratio of excitatory to inhibitory
units of $N_e=800$, $N_i=200$.  However our networks are sparsely
connected.  Our units will be layed out on a ring lattice, with
connectivity $k=10$ connecting each unit to its 10 closest neighbors
(5 clockwise and 5 counterclockwise).  Therefore our networks have
$10^4$ connections out of a possible $\approx10^6$ connections (1\% of
the potential connections).  As we discussed earlier, this sparseness
is more indicitative of biological networks.  Laying out the units in
a ring also serves several purposes.  It embeds the network in a
spatial context, thus giving a meaning to the distance between nodes,
and the length of connections between nodes.  These wiring lengths
will be used in the second simulation to provide a proportional
propagation delay to the signals transmitted as spikes along the
network connections.
\subsection{Research Questions}
\label{sec-1-4}

In the work presented here, our main research question is: how are
local and global synchronization of unit spiking effected by network
architecture?  In particular, does synchronization across a large
network of spiking units appear to be related to the path lengths
available for the transmission of information across the network?
Also, how does clustering, or the cliquishness of neighbors, affect
spiking dynamics?  In this work, we assume that global synchronization
of spiking behavior is in some sense good, and is necessary for such
crucial tasks in biological brains such as multi-modal integration,
attentional focus and awareness, conscious processing, etc.  In the
models presented in this paper, we therefore make a simplifying
assumption that the ability to form network wide synchronized spiking
behavior indicates a positive capability of the network, and when such
global behavior is not possible, the networks are deficient in
their ability to perform global information processing tasks.

In the following simulations, we use the WS interpolation method in
order to vary network architecture from a completely regular (ring)
lattice, through the small-world region up to a completely random
graph.  We will use the WS interpolation method to form networks of
Izhikevich spiking units \cite{izhikevich-2003}, connected with the
varying properties of $L$ and $C$ as we simulate networks transitioning
though this parameter space.  We show that spiking synchronization
dynamics are highly dependent on $L$ and $C$ parameters, and have a phase
transition in the same small-world region from an inability to form
global synchronization, to a strong ability to form global
synchronized behavior.

We further investigate in these simulations, the question of what role
wiring length or cost may play in the formation of global synchronized
behavior.  In particular, we investigate the cost of wiring length in
the form of a more realistic propagation delay of spiking signal
information.  Most simulations of spiking networks are performed
without taking into account the real cost of wiring or path length in
terms of the dynamical behavior.  In this paper, we show that wiring
length, in the form of such propagation delays, does interfere with
the ability of a network to form global synchronization, as one would
expect.  Thus global synchronization may also be constrained by the
average path length $L$ needed to propagate information, which gives
another reason why completely random networks may be worse, in an
information processing sense, than networks in the small-world region
for doing information processing tasks that require global or at least
more large-scale regional synchronization in order to properly
function.

In the following sections, we will first present our general methods
and models used in the simulations in this paper.  We will next
discuss the specifics of two experiments, illustrating the effects of
small-world network architecture on global spiking synchronization
behavior, both without and with propagation delay of spike firings.
We conclude with a more general discussion of the simulation results,
and conclusions and ideas for future investigations.
\section{Method}
\label{sec-2}
\subsection{Izhikevich Spiking Network Model}
\label{sec-2-1}

In all of the simulations\footnote{All simulations, scripts and data analysis artifacts used in
this research report can be found at and reproduced from
\href{https://bitbucket.org/dharter/sync-dyn-small-world-net}{https://bitbucket.org/dharter/sync-dyn-small-world-net}.
 } presented in this work, we use standard
Izhikevich units, as first described in \cite{izhikevich-2003}.  In that original
work, a spiking model was presented that is both computationally
efficient, but capable of displaying a large range of spiking dynamics
similar to those seen in real biological spiking networks.   Izhikevich
units define only two variables to simulate the activity of a unit, 
the membrane potential $v$ and a membrane recover variable $u$, with
an auxiliary after-spike resetting rule:

\begin{equation}
v' = 0.04v^2 + 5v + 140 - u + I
\end{equation}

\begin{equation}
u' = a(bv - u)
\end{equation}

\begin{equation}
\textrm{if} \quad v \ge 30 \textrm{mV, then}
\left\{
\begin{array}{l}
v \leftarrow c \\
u \leftarrow u + d \\
\end{array}
\right.
\end{equation}

The $a,b,c,d$ parameters can be set to produce different spiking
dynamics.  $I$ represents input to the unit, which comes from a
combination of randomly generated thalmic background input, plus input
received from connected units generated from their spike firings.  We
used two sets of these parameters, one for excitatory units and one
for inhibitory units, the same as in the original Izhikevich example.

In the \cite{izhikevich-2003} model description, Izhikevich showed an
example of synchronized spiking behavior in a fully connected network
of $N=1000$ units.  Our simulations also use $N=1000$, but in sparsely
connected networks with varying properties connected as ring lattices
(see next section).  We also use the same mix of two types of basic
units, excitatory ($N_e=800$) and inhibitory ($N_i=200$).  Excitatory
and inhibitory weights, as well as simulated random thalmic input,
were chosen in order to best illustrate the changing spiking
synchronization dynamics.  However, it appears that the dynamics
reported in this research are robust over a wide range of parameters
in the weight and thalmic stimulation space.  All weights and external
thalmic input parameters are generated randomly, using a uniform
distribution on the open interval $[0.0, 1.0)$, multiplied by the
indicated parameter value. Table \ref{tbl:izhikevich-net-params} gives
a summary of all of the important parameters, and their settings used
in all Izhikevich spiking networks presented in this research.

\begin{table}[htb]
\caption{Parameters used for all Izhikevich network simulations.} \label{tbl:izhikevich-net-params}
\begin{center}
\begin{tabular}{lp{3in}r}
 Parameter  &  Description                                                    &  Value  \\
\hline
 $N$        &  Total units in simulation network                              &   1000  \\
 $N_e$      &  Number of excitation units in network                          &    800  \\
 $N_i$      &  Number of inhibitory units                                     &    200  \\
 $w_e$      &  The excitatory weight scaling factor                           &     32  \\
 $w_i$      &  The inhibitory weight scaling factor                           &     22  \\
 $t_e$      &  The amount of random thalmic stimulation for excitatory units  &      3  \\
 $t_i$      &  The amount of random thalmic stimulation for inhibitory units  &     11  \\
\end{tabular}
\end{center}
\end{table}
\subsection{Small-World Interpolation}
\label{sec-2-2}

\begin{figure}[htb]
\centering
\includegraphics[width=.9\linewidth]{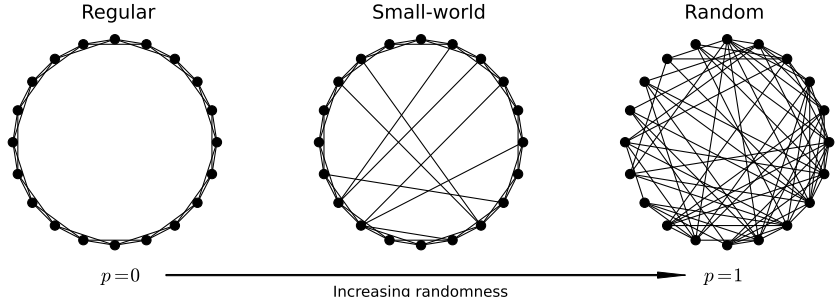}
\caption{\label{fig:ws-interpolated-networks}WS interpolated network results.  The example ring lattices shown here have $N=20$ units and $k=4$ connectivity.  $p$ is the probability of rewiring a connection during interpolation.}
\end{figure}

We organized the spiking networks as ring lattices, both to give a
meaning of spatial layout and distance associated with wiring lengths,
as well as to reproduce the WS interpolation from regular to random
networks (see Figure \ref{fig:ws-interpolated-networks}).  All
simulations in this paper consisted of an initial ring lattice of
$N=1000$ units.  All units were connected to their $k=10$ nearest
neighbors, as in the original Watts and Strogatz analysis.  Here
nearest neighbor is defined spatially along the ring, so for $k=10$
connectivity, the 5 closest neighbors clockwise and counterclockwise
were connected.  In terms of the propagation of spikes, the generated
networks were treated as directed networks, so while initially
neighbors would be connected with one another, the weight of the
connections can be different from source to destination unit and back
from destination to source.  Likewise, random rewiring only affected
one directed connection at a time, so two units connected to one
another might not be reciprocally connected after random rewiring.
All networks thus had $10,000$ total connections in the network (
$k=10 \times N=1000$ ).  Also note that this means that only a total
of $1\%$ of the possible connections are present when compared to a
fully connected network ( $10^4$ actual connections of $10^6$
potential connections).

Interpolation from regular lattice to random network was recreated as
described in \cite{watts-strogatz-1998}.  The basic algorithm is as
follows:

\begin{enumerate}
\item Create a regular ring lattice of size $N$ nodes and connectivity $k$.
\item Examine every (directed) edge.  With probability $p$ reconnect the
   edge to a new randomly selected destination.  The destination is
   selected among all nodes (except the source) with uniform
   probability.
\end{enumerate}

When $p=0$, no rewiring occurs, and we end up with the same regular
ring lattice we started with (Figure
\ref{fig:ws-interpolated-networks}, left).  When $p=1.0$, every
connection is rewired and the result is a randomly connected network
(Figure \ref{fig:ws-interpolated-networks}, right).  As p increases,
at first a small number of long-range interconnects are formed (Figure
\ref{fig:ws-interpolated-networks}, middle).  This is the beginning of
the small-world region of network organization.  As p increases from
$0.0 \rightarrow 1.0$, the network experiences phase changes at
different locations of $p$ for crucial measurements such as $C$ and
$L$.  Further there are large areas where clustering $C$ is high, as
in a regular lattice, but $L$ is low, indicating short average path
lengths for global communication. This area of network connectivity
with high $C$ but low $L$ is known as the small-world region (see
figure \ref{fig:sim1-CLS-plt}).
\subsection{General Network Spiking Dynamics}
\label{sec-2-3}

\begin{figure}[htb]
\centering
\includegraphics[width=.8\linewidth]{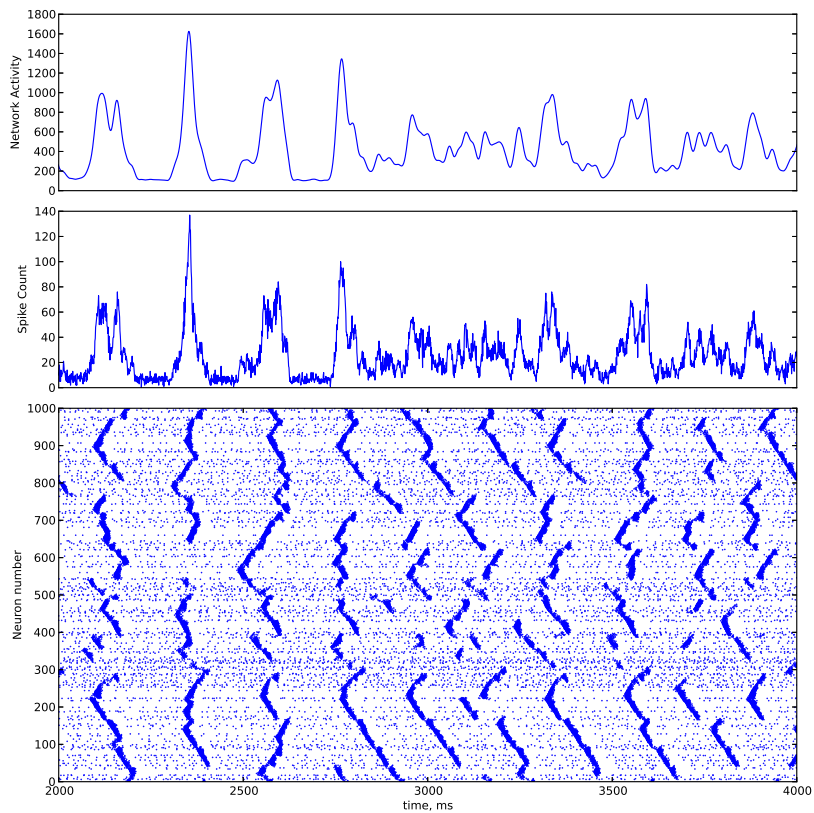}
\caption{\label{fig:raster-net-plt}Spike (scatter) plot (bottom), simple count of spikes at each time step (middle), and convoluted summary of overall mean field spiking activity (top).}
\end{figure}

Figure \ref{fig:raster-net-plt} shows a typical example of network
activity obtained with networks structured as described previously.
This figure was produced with $p=0.001$, so within the small-world
region, but still fairly regular with relatively large average path
lengths $L$.  In this figure, we show a standard raster plot of spike
firings at the bottom, and a summary of spike counts in the middle
part of the figure.  Both plots show activity for $2000ms$ of
simulated time. Raster plots are simply obtained by plotting a point
for every unit, at every time step where the unit is considered to
have fired a spike.  For Izhikevich units, this is when $v>30$,
e.g. when the unit voltage exceeds its spike reset threshold.  The middle
part of Figure \ref{fig:raster-net-plt} shows a crude measure of overall
network activity.  This measure is obtained simply by counting up all
of the spikes that occur at each discrete simulated time step.

Overall this figure demonstrates a few properties of the spiking
dynamics of our ring based regular lattices composed of Izhikevich
spiking units.  For completely regular lattices, at the weights and
thalmic stimulation we are using, highly clustered $C$ networks show a
tendency to produce ripples or waves of spikes that travel from
initiator spots around the ring.  Overall coherence and
synchronization is not possible for $p=0$ and low $p$ networks,
because spike waves must travel in many hops to propagate around the
ring.  The overall measure of network activity shows some
oscillations, but clearly indicates a lack of organized global
structure.  As $p$ increases, however, global synchronization is
enhanced, as we will see when interpolate the networks through the
range of $p$ values.
\subsection{Enhancing Detection of Synchronized Behavior}
\label{sec-2-4}

We found that a simple count of spike activity was a poor measure of
global network synchronization for a couple of reasons.  When used to
estimate the frequency and power of coherent network wide synchronized
waves, simple spike activity hides much of the co-activation
information present.  Spikes that occur within a few, or even just 1,
ms apart do not contribute to a measure of synchronization.  

We enhanced the detection of synchronized firings by convolving each
spike train with a convolution kernel:
\begin{equation}
y = e^{ - (\frac{x}{10})^2}
\end{equation} We used a window width of $30ms$ for our convolution
function.  This has the effect, when summing up spiking behavior, to
allows spikes within a $\pm 15ms$ window to contribute at least some
to the measure of synchronous firing.  Of course, because of the
function we convolved the spikes with, signals $15ms$ apart contribute
much less to each other than when firing at the same time, but still
in general this helps to detect overall synchronized activity, even if
it is spread over a window.  In our tests, the size of the window did
not significantly alter the results we report, with convolution
windows from as small as 6 to as large as 200.  After convolving all
spike trains, we sum up the convolved activity of all units to get a
measure of network activity over time.  Figure
\ref{fig:raster-net-plt} top, from the previous section, displays the
overall network activity measured using the convolved kernel window.
As you can observe, the effect is to significantly smooth the
activity, which also helps in analyzing frequency components.  This is
effectively calculating the local field potential of the simulated
network.

In this research, we introduce a measure of global synchronized
network behavior called $S$, based on the overall convolved spiking
activity.  We define $S$ as the power of the dominant frequency of the
convolved network activity.  $S$ is obtained through a standard power
spectrum analysis, using the fast Fourier transform, from which we
determine the frequency at which the maximum power component was
present in the signal.  For Figure \ref{fig:raster-net-plt}, we
obtained a $S=3.5e^{11}$ power signal at a frequency of $6.8Hz$.
\section{Experiments}
\label{sec-3}
\subsection{Simulation 1: Network Synchronization Dynamics}
\label{sec-3-1}

In simulation 1, we will look at how spike synchronization dynamics
are affected as $p$, the probability of rewiring, is interpolated
through the small-world region, from completely regular to completely
random networks.  We explore the space of $p$ values so that we can
see how behavior changes as a function of $p$.  For each $p$, we run
$NUM\_SIMS=100$ simulations and report the average value of the
measured properties.

\begin{figure}[htb]
 \centering
 \includegraphics[width=.9\linewidth]{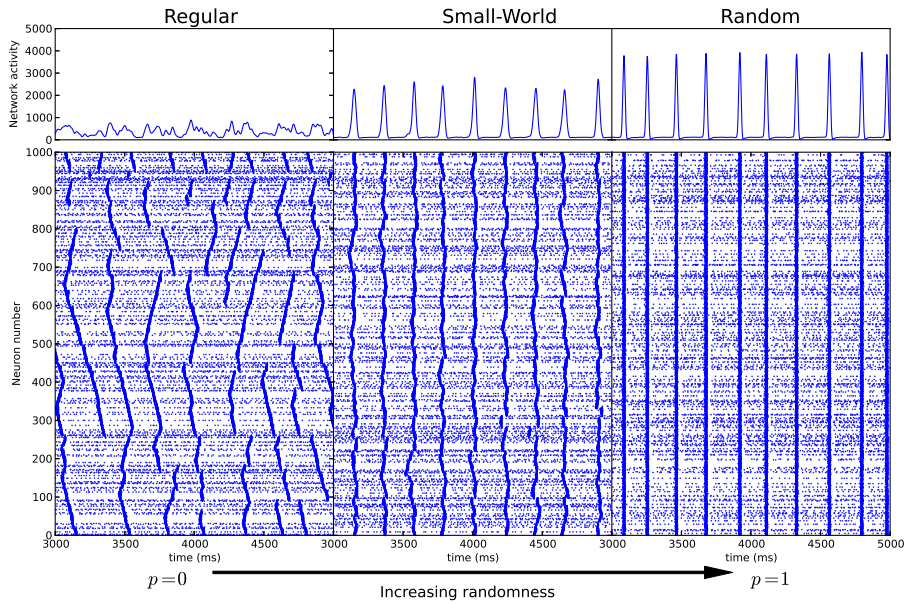}
 \caption{\label{fig:sim1-ex}Typical network/unit dynamics of simulation 1.  Left, $p=0$ regular lattice.  Middle, $p=0.02$ small-world. Right, $p=1$ random network. Top, convoluted network activity.  Bottom, unit spiking activity.}
 \end{figure}

First we look at typical spiking and network activity in three regions
of the $p$ space for our simulations.  In Figure \ref{fig:sim1-ex}, we
show spiking and network activity of 3 separate networks, $p=0.0,
p=0.02, p=1.0$.  As in the previous figures, the top part of the
figure shows the overall convolved network activity, or local field
potential of the network, while the bottom shows a spike raster plot
of all of the spiking activity of all of the units for the selected
time period.  Regular lattice dynamics are typically realized as waves
of activity propagating around the ring, as seen in the figure, left.
Since long range connections are not present, by necessity information
such as a wave of spikes must propagate from unit to unit, cluster to
cluster.  As we increase $p$, more and more long range connections are
formed, shortening overall path lengths.  The middle portion of Figure
\ref{fig:sim1-ex} shows activity for a network firmly within the
small-world region of the $p$ parameter space.  With only $2\%$ of
connections rewired, we see significantly more ability for spikes to
synchronize globally.  Here, already, looking at the global network
activity it is easy to identify global synchronized spike waves,
occurring at a regular frequency.  On the right side of Figure
\ref{fig:sim1-ex} we show activity of the random network, where
$p=1.0$.  Random networks are characterized by connections that jump
across the ring, so synchronized activity is easily and readily
achievable.  Here all units regularly fire in bursts within very small
windows.  Frequencies achieved for global synchronization range from
about \$9Hz.\$ to $15Hz$ typically. The characteristic frequency appears
to be a function of the network size for our simulations.

\begin{figure}[htb]
\centering
\includegraphics[width=.9\linewidth]{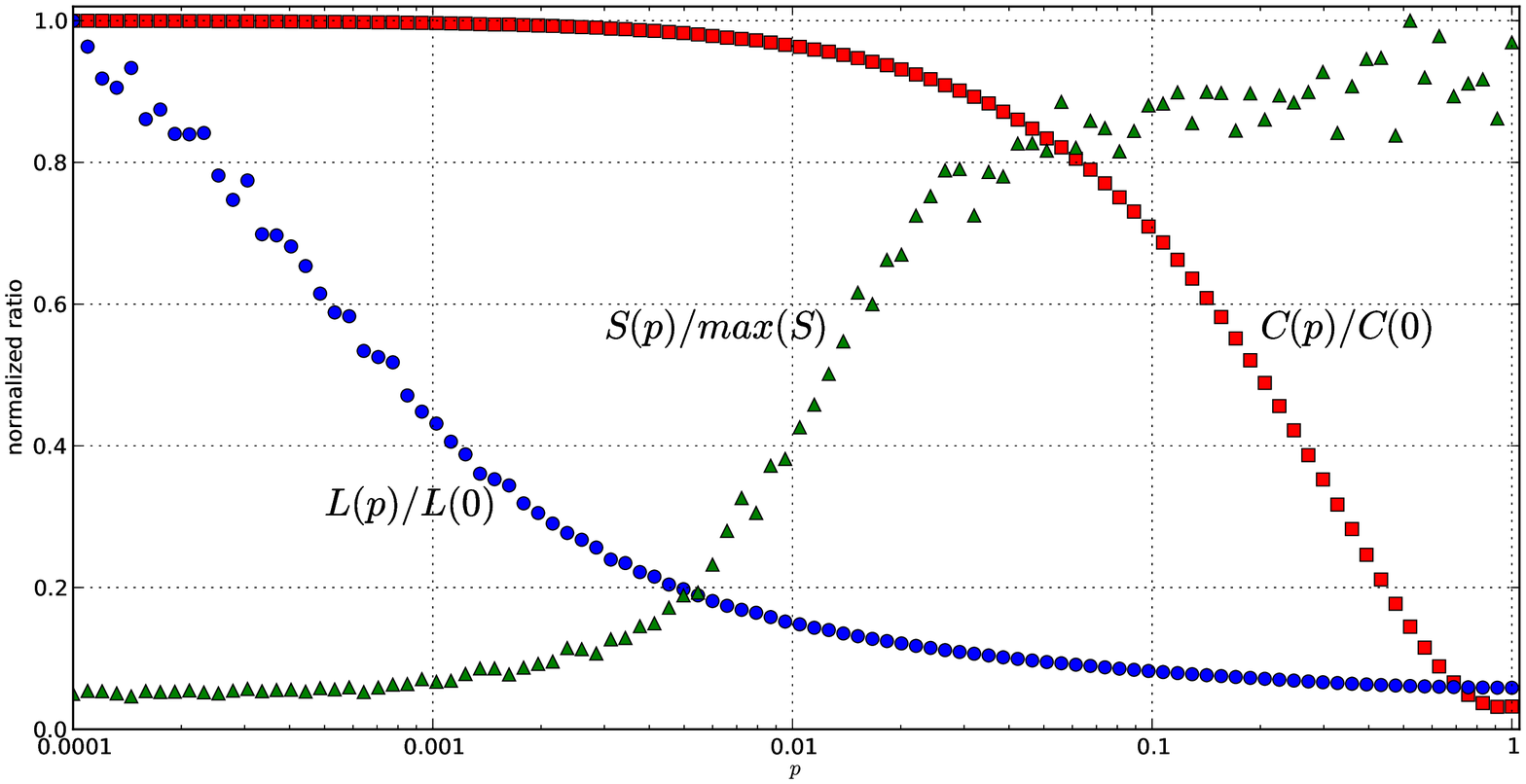}
\caption{\label{fig:sim1-CLS-plt}Simulation 1 results, comparison of $S$ network synchronization activity measure, against $C$ (average clustering) and $L$ (average path length).}
\end{figure}

In order to measure overall synchronization, we ran simulations across
the spectrum of $p$ rewiring probabilities and measured the previously
defined network synchronization parameter $S$.  We ran $NUM\_SIMS=100$
simulations for each $p$ value and averaged the result of all
simulations for the same $p$.  The results of this set of simulations
is presented in Figure \ref{fig:sim1-CLS-plt}.  In this figure we
normalized $S$ by dividing all results by the largest recorded
$max(S)$ over all simulations, thus placing all values in the range
$[0, 1.0]$.  We did this so that we could compare directly with the
original Watts and Strogatz results of $C$ clustering and $L$ average
path length, which we measured for our networks as well in order to
reproduce the basic example of small-world network structure, and
which were originally normalized in this same fashion.  Also note that
the figure use a $log_10$ scale for the $p$ axis.  This is done so
that we can more easily see the phase transitions that occur in the
measured network properties.  Also note that this implies that much of
the phase transition, especially for the $L$ and our $S$ measure, is
occurring for farily minimally wired networks, since $p$ is somewhere
in the range of $0.01 \rightarrow 0.08$ or with only around $1\%$ to
$8\%$ of connections being rewired.  This shows that the introduction
of a small number of long-range interconnections can have major
effects.  And that such networks can still be close to minimally
wired, and thus have fairly low material resource requirements.

As shown in Figure \ref{fig:sim1-CLS-plt}, our network simulations do
reproduce the phase transitions of $C$ and $L$.  Network
synchronization $S$ (triangles) also shows a phase transition in the
critical small-world region.  For $S$ the phase transition is
happening between where $L$ and $C$ occur in the $p$ space.  So $S$
does not begin to show much increase until $L$ has mostly
transitioned.  However $S$ has gone through most of its transition by
the time $C$ begins its phase change.

$S$ does not appear to be directly tied to the phase change of either
$C$ or $L$ alone, but it is more directly related to and sensitive of
changes in overall average path length.  Also note that high $S$
synchronization can be achieved only for fairly low $L$.  But high
global synchronization can be seen for networks either with high
clustering $C$ (around $p=0.02$) or low clustering ($p$ from $0.2
\rightarrow 1.0$).  This brings up an interesting question: is global
synchronization unaffected or unrelated to the clustering of units
in the network?  This observation motivates the question for the next
simulation: what properties might be present that lessen the ability
for networks to synchronize for larger $p$?
\subsection{Simulation 2: Effects of Propagation Delay on Synchronization}
\label{sec-3-2}

One obvious deficiency for studying global network synchronization is
that the time to propagate signals is typically ignored. In many
simulations of network dynamics, there is no real notion of the
spatial layout of units, nor the cost of wiring up such units in terms
of material or delay of propagation of signals.  In the second
simulation, we added a propagation cost proportional to the wiring
length of the connection carrying a spike signal.  Distance is simply
measured as the number of units around the ring (in shortest
direction) one needs to go to get from the source to the destination
of a spike signal.  We added a buffer mechanism to the basic
Izhikevich simulation, so that spike signals could be delivered to
their destination with varying delays depending on their distance from
source.

We found that networks with maximum delays of $20$ were sufficient to
demonstrate the effect, delay costs less than $10$ usually began to
show no effects on our simulations.  For the simulations that follow,
we used a distance scale of 25, meaning that units only 25 or less
distance away on the ring experience a 0 propagation delay, while
units from 26-50 distance away experienced a 1ms delay, up to 19ms
propagation delay for units 451-500 distance on the ring.  Therefore
units have delays in propagating their spike firing inputs ranging
from $0 \rightarrow 19$.  This adds an additional burden on highly
random networks, that have a large number of long range
interconnections.  In such networks, global activity is stifled by the
time that it takes for signals to propagate over the long
range connections.  This, we believed, would have a similar effect to
the propagation effects encountered by regular networks at the $p=0$
end of the space, that also experience propagation delays because of
the need for multiple hops for information to propagate.

\begin{figure}[htb]
\centering
\includegraphics[width=.9\linewidth]{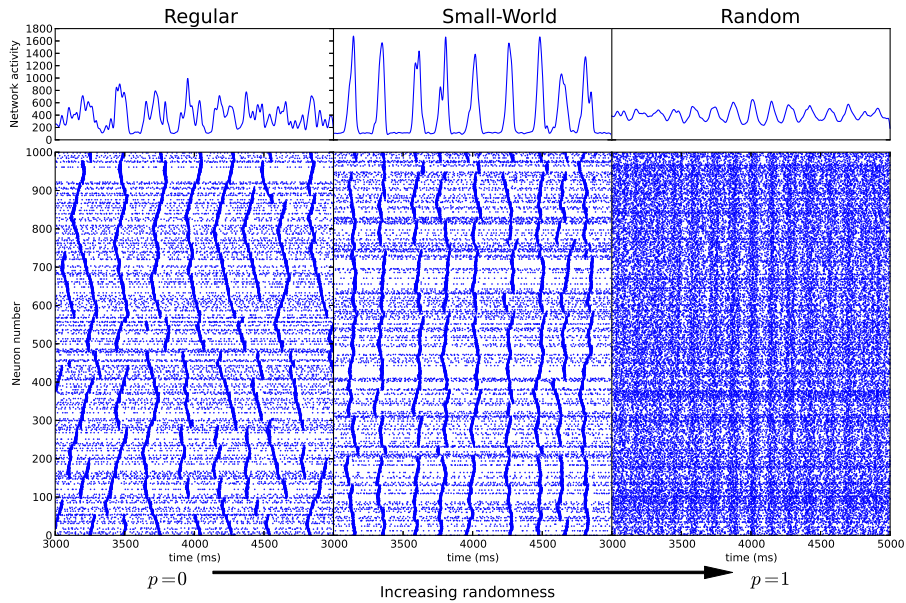}
\caption{\label{fig:sim2-ex}Example network dynamics of simulation 2.  Left, regular lattice $p=0$.  Middle, small-world network $p=0.02$.  Right, random network $p=1$.  As before, the top part of figure depicts overall convolved network activity, while the bottom shows individual unit spiking activity.}
\end{figure}

Figure \ref{fig:sim2-ex} shows typical dynamics of networks in
simulation 2.  Propagation delays do not have much of an effect on
regular networks nor networks with a small number of long range
connections (we again show networks with $p=0$, $0.02$ and $1.0$ random
rewiring).  Long range connections in the small-world region, even
with some delay, are still sufficient to help encourage global
synchronization.  However, the effects on high random $p$ networks is
a complete breakdown of coherent behavior (Figure \ref{fig:sim2-ex},
right).  Here we can definitively see the effect that a lack of
clustering cliquishness has on global network activity.  With neither
well organized neighborhood clusters, nor the ability to effectively
and efficiently transfer information long distances, all coherent
activity disappears from these networks.

\begin{figure}[htb]
\centering
\includegraphics[width=.9\linewidth]{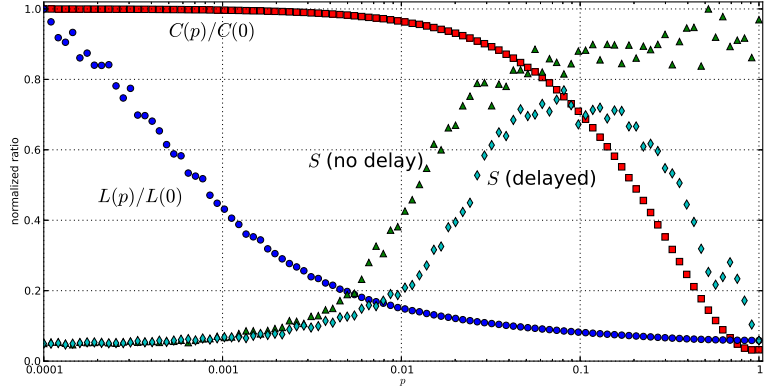}
\caption{\label{fig:sim2-CLS-plt}Simulation 2 results, comparison of $S$ with no propagation delay (triangles) to $S$ with propagation delay (diamonds).}
\end{figure}

As in simulation 1, we wanted to examine the overall synchronization
$S$ of networks with propagation delays.  We again ran $NUM\_SIMS=100$
simulations for each $p$ value reported.  We reproduce all of the 3
$C$, $L$ and $S$ curves from the previous simulation, and impose the
new $S$ obtained on networks with propagation delay.  The results are
shown in Figure \ref{fig:sim2-CLS-plt}.

With propagation delay in the network activity simulation, $S$ still
shows a phase transition in the small-world region, until a certain
point (diamonds).  At some point beyond $p=0.1$, a combination of a
lack of local cohesion from neighborhoods, coupled with insurmountable
propagation delays in passing signals around the network, causes all
coherent activity to cease.  The breakdown of the delayed $S$
synchronization performance appears to be related to the loss of
clustering $C$ of the networks.  This makes sense, without delays we
believe that the whole network in a sense forms a single large
cohesive cluster, thus random networks are able to sustain global
firing without small local neighborhood cluster.  But with delays the
loss of local neighborhood cluster to synchronize activity results in
the loss of any overall coherent activity.  This can be seen when
comparing the dynamics in the simulation 2 random networks, to regular
network dynamics, which while not globally synchronized, do display
wavelets of coherent spike trains that propagate around the ring
spatially.  This plot does show a limitation of our $S$ measure, as
there is a clear difference in activity from the $p=0$ end of the
spectrum to the $p=1$ end of the spectrum (see previous Figure
\ref{fig:sim2-ex}).  At $p=0$ there is still group synchronization,
but such waves are unable to organize into global patterns.  While at
$p=1$ There is no synchronization of any kind within the network, all
activity is random firing of units.
\section{Discussion}
\label{sec-4}

The presence of a phase transition of the global synchronization
dynamics is a nice illustration that the effects a few long-range
interconnections can have on global information transfer.  In this
research, we showed that spiking dynamics undergo such a phase
transition with even a small number of long-range interconnections.
But decreasing path lengths may not be the only factor in this
transition.  When other types of constraints are taken into account,
for example the cost of propagating the signals over longer distances,
we see that large numbers of long-range interconnections may be
suboptimal in more than just material costs and constraints sense.
The effect of transmission delays on global synchronization can only
be overcome with networks in the small-world region, that balance just
enough long range connections to allow for efficient global
communication, but not so much that delayed signals can't properly be
correlated with one another by the network groups.  Further,
clustering may play a role in the ability of small-world networks to
produce global synchronization.  As long as local clustered groups are
present when there are propagation delays, the network appears to be
able to sustain coherent activity.  With delays, if there are coherent
groups then at least some coherent activity and information transfer
is possible.  But if there are no groups, randomness and incoherent
activity will be the norm.

The presence of waves spreading out from a seed point, seen for
regular networks and small-world networks that are sub-threshold for
global synchronization, remind us a bit of Freeman's theory of
collective synaptic synchronization, known as Freeman's theory of mass
action \cite{freeman-2008,freeman-1975}.  Freeman has demonstrated the
presence of wavelets of activity in Brain EEG's, that seed from points
in the neural tissue and spread outwards in synchronized mass action
waves.  These are different from regular synchronized oscillations,
such as gamma and alpha waves.  We speculate that the clustering of
small-world networks can provide seed points for such wavelets of
activity, and that sub-threshold synchronization allows for such seeds
to propagate from group to group, rather than completely dominating
global dynamics.

In this work we do assume global synchronization is an important
property for information processing in dynamical spiking networks.
This of course is not really true, too much synchronization is known
to be a bad property, and can be signs of a seizure or other
malfunction in the functional dynamics.  It does appear that network
dynamics need to be tuned towards organizations where they are close
to, and able to sustain global waves of synchronization for short
periods, but that mostly synchronize in more local groups.
Small-world organization, with highly modular clustering, does appear
to support this type of functional dynamic readily.

As we mentioned in the discussion of the experiment results, $S$ is a
crude measure of network synchronization.  It certainly is incapable
of detecting the wavelet patterns that high $C$ neighborhoods seems to
produce.  In addition it is not able to determine if there are other
frequency dynamics present in the networks, though of course we could
always also examine the second, third, etc.  dominant frequencies that
are not harmonics of each other.  Other measures of global
organization, that don't depend on a local summed field potential,
might be interesting to apply to these ring lattice organized
networks.  Perhaps measures of synchronized spiking activity based on a nodes
friendship graph (immediate neighbor, 2nd neighbors, etc.)  might be
of interest.  For the case where connections have a propagation delay,
in effect a distance weight (different from the action potential
weight), we might want to examine groups based on the time of travel of
signals on the shortest distance path between units.

In this research report we have concentrated solely on small-world
network organization and its effects on spiking dynamics.  Most
complex real world networks, including brain networks, have properties
that are not captured by the WS small-world interpolation.  In
particular, brain and other real networks show scale-free degree
distributions, where the degree of a node is simply the count of the
number of connections to or from the node \cite{albert-barabasi-2002}.
For WS interpolated networks, degree distributions are fairly uniform
at all points along the $p$ spectrum.  For scale-free degree
distributions, there are present a few hub nodes that are highly
connected, while the majority of nodes have very few connections.  A
similar analysis on the effects of interpolating between small-world
networks with uniform degrees (WS small networks) to small-world
networks with scale-free degree distributions would be an interesting
follow up to this work.  The presence of hub nodes in such networks
provide short range paths, allowing for efficient global
communication.  However, clustering can be increased and decreased in
such scale-free networks, to tune the amount of sub to critical
threshold wavelet generation and global synchronization.
\section{Conclusion}
\label{sec-5}

In the reported research, we showed the results of interpolating network
structure from regular, through small-world regimes to random networks, on the
functional dynamics of spiking networks. Functional dynamics of sparsely, but strongly
connected spiking units depend critically on the network organization.  Regular
lattices are characterized by packets of spreading spike wavelets, that originate at 
a point and spread around the ring lattice.  Without long-range connections, global
synchronized behavior is not possible.  At the other extreme, random networks
are easily and strongly globally synchronized.  Small-world networks begin
to allow global organization rather quickly.  Within the small-world region,
larger groups become coherently connected with a few long-range connections, and
global synchronized behavior emerges.  We demonstrated that this emergence
of global synchronization ($S$) takes the form of a phase transition in the small-world
region, similar to and between the phase transitions of clustering ($C$) and
average path length ($L$).

In addition to exploring the functional dynamics of small-world networks, we looked
at the effect of adding more realistic propagation delay constraints to the
functional dynamics of such spatially constrained networks.  Adding propagation
effects of spike signals makes it impossible for random networks to sustain
global synchronization behavior (once delays are sufficient).  Only networks
in the small-world region are able to successfully sustain global synchronized
behavior when significant propagation delays are present.  This is due to the
presence of a small number of long-range paths that allow for information
to be transmitted over the network globally, as well as the reinforcement of
synchronized behavior through the tight coupling of neighborhood clusters.
We showed that the phase transition in global synchronization breaks down
under the presence of propagation delays, apparently as a function of the
decreasing clustering in the network as they become more random.

\section*{Acknowledgments} This work was partially supported by NSF
grant \#0916749 and DOE grant \#DE-SC0001132.  Simulation resources and
time were provided by the Texas A\&M - Commerce Lion HPC computing
cluster.

\bibliographystyle{hplain}
\bibliography{synchronization-dynamics-small-world}

\end{document}